\documentclass[final,5p,times,twocolumn]{elsarticle}

\usepackage{amsmath}
\usepackage{mathtools}
\usepackage{xcolor}
\usepackage{enumitem}
\usepackage{comment}
\usepackage{graphicx}
\usepackage{multirow}
\usepackage{array}
\usepackage{caption}
\usepackage{fontawesome5}
\usepackage{hyperref}
\usepackage{wrapfig}
\usepackage{longtable}
\usepackage{times}
\usepackage{latexsym}
\usepackage{geometry}
\usepackage{array}
\usepackage{latexsym}
\usepackage{color, colortbl}
\usepackage{booktabs}
\usepackage{adjustbox}

\usepackage{amssymb}

\makeatletter
\def\ps@pprintTitle{%
 \let\@oddhead\@empty
 \let\@evenhead\@empty
 \let\@oddfoot\@empty
 \let\@evenfoot\@empty}
\makeatother

\begin{document}

\begin{frontmatter}



\title{A Unified Framework with Novel Metrics for Evaluating the Effectiveness of XAI Techniques in LLMs}

\author[1]{Melkamu Abay Mersha} 
\author[2]{Mesay Gemeda Yigezu}
\author[1]{Hassan Shakil} 
\author[1]{Ali K. AlShami }
\author[1]{Sanghyun Byun}
\author[1]{Jugal Kalita}


\affiliation[1]{organization={College of Engineering and Applied Science, University of Colorado Colorado Springs},
            postcode={80918},
            state={CO},
            country={USA}}

\affiliation[2]{organization={Instituto Politécnico Nacional (IPN), Centro de Investigación en Computación (CIC)},
            postcode={07738},
            state={Mexico city},
            country={Mexico}}
            

            







\begin{abstract}
The increasing complexity of LLMs presents significant challenges to their transparency and interpretability, necessitating the use of eXplainable AI (XAI) techniques to enhance trustworthiness and usability. This study introduces a comprehensive evaluation framework with four novel metrics for assessing the effectiveness of five XAI techniques across five LLMs and two downstream tasks. We apply this framework to evaluate several XAI techniques—LIME, SHAP, Integrated Gradients, Layer-wise Relevance Propagation (LRP), and Attention Mechanism Visualization (AMV)—using the IMDB Movie Reviews and Tweet Sentiment Extraction datasets. The evaluation focuses on four key metrics: Human-reasoning Agreement (HA), Robustness, Consistency, and Contrastivity. Our results show that LIME consistently achieves high scores across multiple LLMs and evaluation metrics, while AMV demonstrates superior Robustness and near-perfect Consistency. LRP excels in Contrastivity, particularly with more complex models. Our findings provide valuable insights into the strengths and limitations of different XAI methods, offering guidance for developing and selecting appropriate XAI techniques for LLMs.

\end{abstract}


\end{frontmatter}



\section{Introduction}

\label{section:introduction}

With AI systems becoming integral to decision-making processes across safety-critical sectors such as healthcare, autonomous driving, and finance, ensuring transparency in these systems is essential. Large language models (LLMs), such as GPT \cite{radford2018improving, shakil2024evaluating} and BERT \cite{devlin2018bert}, play a pivotal role in enhancing the capabilities of modern AI through advances in natural language processing tasks \cite{mersha2024semantic}. However, these models' inherent complexity and opacity make it essential to gain a deeper understanding of the underlying principles driving their decision-making processes.

Explanations of LLMs' decision-making processes have become essential following their revolution and deployment across broader applications, helping to uncover potential flaws or biases \cite{das2020opportunities}, build user trust \cite{arrieta2020explainable}, facilitate regulatory compliance \cite{regulation2016regulation}, and guide the responsible integration of AI models into diverse sectors \cite{langer2021we}.

Explainable Artificial Intelligence (XAI) techniques aim to bridge the gap between complex AI models and human interpretability by providing insights into how a model arrives at a particular decision \cite{shah2021neural, mersha2024explainability}. Post-hoc XAI techniques, which explain models after training, can be categorized into five groups: model simplification \cite{ribeiro2016should}, perturbation-based methods \cite{zeiler2014visualizing}, gradient-based approaches \cite{sundararajan2017axiomatic}, Layer-wise Relevance Propagation (LRP) \cite{bach2015pixel}, and attention mechanisms \cite{honnibal2017spacy}. There are a variety of XAI techniques within each category.  However, there is no standardized framework for selecting the most suitable technique for a specific model or systematically evaluating the effectiveness of each method.


Most XAI methods often produce inconsistent and sometimes contradictory explanations for LLMs, complicating the process of validating their reliability \cite{hassija2024interpreting}. This inconsistency arises due to variations in LLM's behavior and a lack of standardized criteria for assessing explanation quality \cite{pawlicki2024evaluating}. Moreover, different methods may be more suited to specific model architectures or application domains, further complicating their selection and evaluation \cite{atanasova2024diagnostic}. Current evaluation approaches rely on feature saliency scores, which may not always capture the nuances of a model’s reasoning process, potentially leading to misleading conclusions when different features have identical saliency scores. 

To address these challenges, we propose a comprehensive evaluation framework that integrates four text ranking-based metrics along with saliency scores.  The proposed metrics evaluate XAI explanations from different perspectives: Human-reasoning agreement measures how well the explanations align with human rationales, Robustness tests the stability of explanations under data inconsistencies, Consistency examines their alignment with the model’s internal attention patterns, and Contrastivity assesses how well they differentiate between distinct classes (e.g, positive vs negative classes).

Our approach considers critical factors such as model complexity, language diversity, downstream task variations, and application-specific requirements, ensuring a holistic evaluation across different scenarios. Unlike existing frameworks that focus on isolated aspects, our framework provides a more comprehensive evaluation, making it adaptable to diverse model architectures and use cases.

We apply this framework to evaluate five distinct XAI categories on five LLMs of varying complexities, using two text classification tasks (short and long text inputs). By integrating these metrics, we offer a nuanced understanding of the strengths and limitations of each XAI technique, enabling more informed selection and use of XAI methods for specific models and applications.




The contributions of this study are: 
\begin{itemize} 
\setlength{\itemsep}{0pt}
  \setlength{\parskip}{0pt}
    \item We provide comparative analyses to guide the selection of suitable XAI techniques for different LLMs in downstream tasks. 
    \item We propose a comprehensive end-to-end evaluation framework with four evaluation metrics to assess the effectiveness of XAI techniques across various LLMs. 
    \item We analyze and compare five XAI categories across five LLMs using these metrics on two downstream tasks. 
    \item We compare and contrast explanations with human rationales to assess alignment with LLMs' decision-making processes. 
\end{itemize}

\begin{figure*}[hbt!]
\centering
{\includegraphics [width=\textwidth]{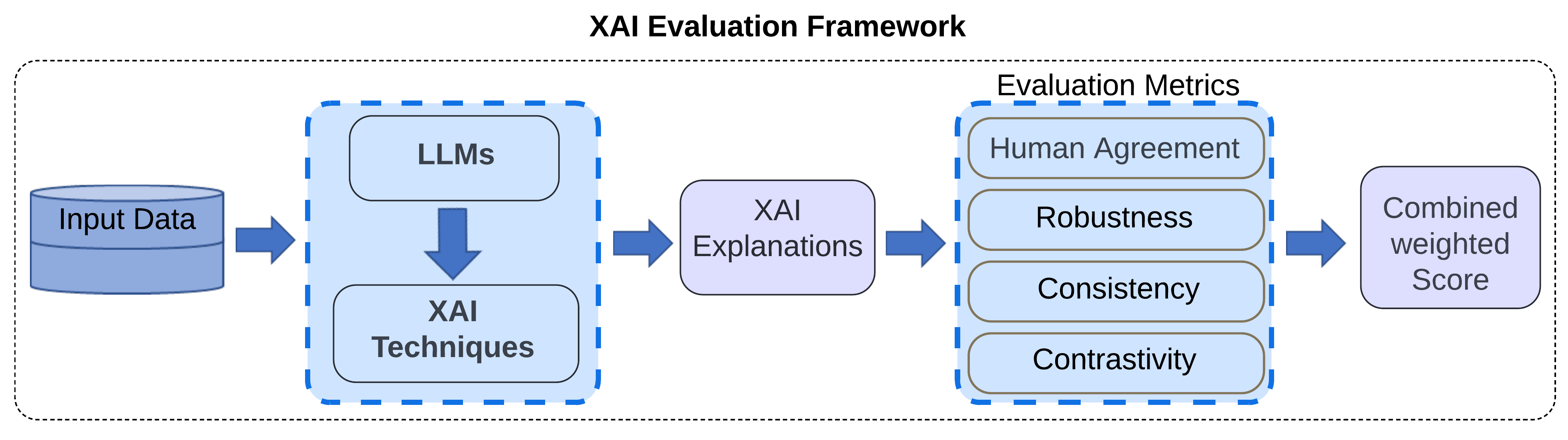}}
\caption{
An overview of our comprehensive XAI evaluation framework for assessing the effectiveness of explainability techniques across different scenarios.} 
\label{fig:Evaluation_framework}

\end{figure*}

\section{Related Work}  \label{section}

The black-box nature of large language models (LLMs) raises concerns about explainability and transparency. XAI techniques aim to reveal the decision-making processes of these models, enhancing user trust \cite{mersha2024explainable}. However, their effectiveness is often questioned due to potentially misleading or insufficient explanations, challenging the goal of making LLMs truly interpretable and trustworthy. This study examines XAI techniques not specifically designed for LLMs, evaluating their ability to improve interpretability without compromising performance. XAI methods are categorized as \textit{ante-hoc} (applied during training) or \textit{post-hoc} (applied after deployment).

Post-hoc explainability techniques, either model-agnostic or model-specific, can be grouped into five categories: (1) \textit{Model simplification techniques} like LIME \cite{ribeiro2016should} simplify complex models into more interpretable ones; (2) \textit{Perturbation-based techniques} like SHAP \cite{shapley1953value} modify feature values to assess their impact on predictions; (3) \textit{Gradient-based techniques} such as Integrated Gradients \cite{sundararajan2017axiomatic}, explain models by analyzing gradients; (4) \textit{Layer-wise Relevance Propagation (LRP)} methods like basic LRP \cite{montavon2017explaining}, LRP with Taylor Decomposition \cite{chefer2021transformer}, and LRP with Conservative Propagation \cite{ali2022xai} redistribute scores backward through layers; (5) \textit{Attention mechanism techniques} such as Attention Rollout \cite{karras2020analyzing} and Attention Mechanism Visualization \cite{honnibal2017spacy} visualize influential input features by analyzing attention weights.

Due to the diversity of XAI methods, systematic assessment is necessary. Counter-examples reveal weaknesses in explainability techniques \cite{kindermans2019reliability} but do not provide comprehensive evaluation \cite{jacovi2020towards}. Human evaluations often reflect subjective judgments rather than actual effectiveness \cite{bansal2021does}. Alternative methods include Ground Truth Correlation, comparing human-identified salient words with XAI outputs \cite{arras2022clevr}, and sufficiency analysis by removing salient tokens and observing performance changes \cite{deyoung2019eraser}. Other studies have assessed SHAP and LIME on traditional models like random forests and SVMs \cite{arreche2024xai} and evaluated various techniques on CNN, LSTM, and BERT models \cite{atanasova2024diagnostic, mersha2024ethio, shakil2024utilizing}.

Existing evaluations of explainability techniques vary widely with respect to models, methods, and tasks. Some focus on single architectures \cite{deyoung2019eraser}, while others examine multiple models using various techniques \cite{atanasova2024diagnostic}. Many are dataset-specific \cite{guan2019towards}. Additionally, most research overlooks the influence of model complexity on the effectiveness of explainability methods. Transformer models range from millions to billions of parameters \cite{shakil2024abstractive, tonja2023first}. Yet, current research insufficiently explores how explainability methods perform across these complexities or which methods are best suited for different levels of model complexity.

These limitations underscore the need for our comprehensive study, which systematically evaluates the effectiveness of XAI methods across LLMs of varying complexities. By integrating four robust evaluation metrics, we aim to assess how LLM complexity influences XAI performance and to identify the most effective XAI techniques tailored to specific complexities, ensuring optimal application across diverse models and tasks.

\section{Methodology}  
\label{section:methodology}
Figure~\ref{fig:Evaluation_framework} presents a proposed framework designed to evaluate the effectiveness of XAI techniques for LLMs. The framework begins by processing input data through a trained LLM, where various XAI techniques are applied to generate explanations for the model's decisions. These explanations are then assessed using a set of critical evaluation metrics. Each metric provides insights into different aspects of the explanation quality \cite{mersha2025evaluating}.

\subsection{Evaluation Metrics}  
Evaluation metrics for XAI techniques are crucial as they provide quantitative measures to assess the quality and reliability of explanations \cite{nauta2023anecdotal}, \cite{mersha2024explainable}. These metrics ensure that explanations accurately reflect the model’s behavior. They also enable the systematic comparison and improvement of XAI methods. We build a comprehensive evaluation framework to evaluate various XAI categories by adopting and enhancing the existing metrics, and introducing new metrics. 

\subsubsection{Human-reasoning Agreement (HA)}  
The HA metric measures the alignment between XAI explanations and human reasoning \cite{deyoung2019eraser}, \cite{atanasova2024diagnostic}, using human-annotated data as a reference. However, it is important to note that the assumptions regarding the high degree of agreement between the feature importance scores (such as word saliency scores in this study) provided by explainability techniques and those from human-annotated datasets are not always valid. For example, while word saliency scores may be similar, the specific words identified as salient can differ significantly, posing a limitation in using metrics like cosine similarity, Pearson correlation, and intersection-over-union. To address this, we employed token/word ranking and Mean Average Precision (MAP) methods, ranking important tokens based on saliency scores to compute the Average Precision (AP) and MAP for better agreement evaluation.

\textbf{\textit{Mean Average Precision (MAP)}}
MAP provides a comprehensive measure of agreement by evaluating the precision of ranked tokens in XAI explanations relative to human annotations. For a single instance, Average Precision (AP) is calculated as, Equation~\ref{equ.AP}:

\begin{equation} \label{equ.AP} AP = \frac{\sum_{k=1}^{n} (P(k) \times rel(k))}{n} \end{equation} 
 \noindent where $k$ is the rank, $n$ is the number of retrieved words, $P(k) $ is the precision at rank $k$, and $rel(k)$ is 1 if the word at rank $k$ matches human annotations, and 0 otherwise, Equation \ref{equ.relevance}. A high AP indicates strong agreement for that instance.
The Mean Average Precision (MAP) aggregates AP scores over all instances using Equation~\ref{equ.MAP}:

\begin{equation} \label{equ.MAP} MAP = \frac{\sum_{n=1}^{N} AP_n}{N} \end{equation} \noindent where $N$ is the total number of instances. A higher MAP value (closer to 1) signifies consistent agreement between AXI explanations and human rationales across all instances, while a lower value (closer to 0) indicates variability or disagreement.

\subsubsection{Robustness} 
This metric evaluates the robustness of explanations generated by explainability techniques under varying conditions such as input modifications, adversarial perturbations, and model retraining \cite{rosenfeld2021better, nogueira2018stability}. It measures how consistently explanations behave across diverse scenarios, providing insights into the reliability of model interpretations.

We assess the stability of explanations when a slight perturbation is applied to the input instance $X$, resulting in a modified instance $X'$. Traditional robustness measures often compare top-k saliency scores between $X$ and $X'$, which may not fully capture the variations due to differences in rank or score magnitudes \cite{arreche2024xai}. To address this, we employ element-wise differences and averaging techniques to quantify robustness metrics at both the token/word and instance levels based on relevance computation.

\textbf{Relevance Function:} The relevance function, $rel(k)$, serves as a binary indicator to determine the relevance of each word $k$ in a model's decision-making process. If the word $k$ is included in both $X $ and $X'$, it returns 1; otherwise, it returns 0.
 
\begin{equation} 
\label{equ.relevance}
\text{rel}(k) = 
\begin{cases} 
1 & \text{if } k \in X \\
0 & \text{otherwise}
\end{cases}
\end{equation} 
where $k$  is a relevant word to the model's decision-making process.\\
To create a modified instance ($X’$), a perturbation $\delta_{i}$ is applied such that $X’$=$X$ +$\delta_{i}$.  This perturbation $\delta_{i}$ typically involves various techniques such as masking, replacing words with synonyms, removing words, or applying other modifications to words with high or low salience scores.

\textbf{Element-wise Difference $d(k)$:} For each word $k$, the function $d(k)$ precisely quantifies the change in the saliency scores of a word k between $X$ and $X’$, as identified by $rel(k)$. $ d (k) $ is computed using Equation~\ref{equ.elementWise}: 

\begin{equation}
\label{equ.elementWise}
d(k) = \|X[k] - \left ( X'[k] \times {rel}(k) \right) \| 
\end{equation} 
\noindent  where $X[k]$ and $X'[k]$ are the saliency scores for word $k$ in the original and modified inputs, respectively, and $\text{rel}(k)$ is 1 if word $k$ appears in both $X$ and $X'$; otherwise, it is 0.

 
\textbf{Average Difference (AD):} AD aggregates the individual differences $d(k)$ for all relevant words and provides a single metric for each instance. AD reflects the average magnitude of change in an individual explanation due to an input modification, described by Equation~\ref{equ.AD}.

\begin{equation}
\label{equ.AD}
\text{AD} = \frac{1}{K} \sum_{k =1} ^{K} d(k)
\end{equation}
where $K$ represents the total number of words explaining a given instance. A lower AD indicates that explanations remain stable for individual instances under slight input modifications.

\textbf{Mean Average Difference (MAD):} MAD is a dataset-wide metric that averages the AD values across all instances, providing a global measure of the explanations' robustness in response to input data changes throughout the dataset. Mathematically, it is represented by Equations \ref{equ.MAD1} and ~\ref{equ.MAD} 
 
\begin{equation}
\label{equ.MAD1}
\text{MAD} = \frac{1}{N} \sum_{n =1} ^{N} AD
\end{equation}

\begin{equation}
\label{equ.MAD}
MAD = \frac{ \sum_{n=1}^N \left( \frac{1}{K} \sum_{k=1}^K d(k) \right) }{N}
\end{equation}
where $N$ is the total number of instances.\\
Lower AD and MAD scores indicate that the explainability technique performs robustly well both at the instance level and across diverse instances, respectively.

\subsubsection{Consistency (Attention Reasoning)}  
Models with diverse architectures often exhibit lower explanation consistency \cite{huang2022conceptexplainer}. Our primary focus is to examine the similarity of attention-based reasoning mechanisms rather than prediction outputs for models that share the same architecture. To this end, we consider a set of models that are structurally identical but are trained with different random seeds and initialization. This approach enables us to investigate the internal decision-making pathways of these models rather than simply matching output predictions, as different models can produce the same prediction through varied reasoning processes.

Let \( M_{a} \) and \( M_{b} \) denote two models with the same architecture but trained using different random seeds, and \( x_i \) be a specific input instance. We define two types of distances to assess model similarity: Attention Weight Distance \( D_A(M_a, M_b, x_i) \) and Explanation Score Distance \( D_E(M_a, M_b, x_i) \). These distances are calculated using standard similarity metrics like Cosine similarity or Euclidean distance. The attention weight distance helps quantify the difference in how models attend to various input features. In contrast, the explanation score distance reflects the divergence in how explainability techniques describe each model’s predictions.

For the attention weight distance, we compute the average attention weights across multiple layers for a given model. Let \( A_l(M, x_i) \) represent the attention weights at the \( l \)-th layer for an input \( x_i \) in model \( M \). For a model with \( L \) attention layers, we calculate the average attention weights using Equation~\ref{equ.AverageAttention} :

\begin{equation}
\label{equ.AverageAttention}
\overline{A}(M, x_i) = \frac{1}{L} \sum_{l=1}^L A_l(M, x_i). 
\end{equation}

\noindent This average attention weight vector, \( \overline{A}(M, x_i) \), provides a consolidated view of the model’s attention across layers, making it easier to compare the attention mechanisms of different models. The distance between the two models is computed with Equation~\ref{equ.distanMaMb},
 
\begin{equation}
\label{equ.distanMaMb}
D_{A}(M_a, M_b, x_i) = D_{A}(M_a(x_i), M_b(x_i)),
\end{equation} 
capturing the differences or similarities in their attention mechanisms for a given input. $\overline{A}(M_a, x_i)$  and $\overline{A}(M_b, x_i)$ are the averaged attention weights of models $M_a$ and $M_b$ for the input $x_i$, as shown by Equation~\ref{equ.AverageAttention}. Then $D_{A}(M_a, M_b, x_i)$ computed using Equation~\ref{equ.MeanAverageAttention}:
 
\begin{equation}
\label{equ.MeanAverageAttention}
D_{A}(M_a, M_b, x_i) = D_{A}(\overline{A}(M_a, x_i), \overline{A}(M_b, x_i)).
\end{equation}

Similarly, explanation score distance is calculated by comparing the explanation scores generated by a selected explainability technique for the predictions of two models. The distance is defined as Equation~\ref{equ.explantion}
 
\begin{equation}
\label{equ.explantion}
D_{E}(M_a, M_b, x_i) = D(E(M_a, x_i), E(M_b, x_i)),
\end{equation} where \( E(M, x_i) \) denotes the explanation scores for a given input. This distance highlights the extent to which the models’ reasoning mechanisms align in terms of explaining the same prediction, regardless of their output.

To analyze consistency at the instance level, we compare the attention weight distances and explanation score distances for individual inputs. If these distances are similar, it indicates that models with similar attention patterns produce consistent explanations, demonstrating the reliability of the chosen explainability technique. For a broader analysis, we assess consistency across the dataset by computing the Spearman’s rank correlation coefficient (\( \rho \)) between attention weight distances and explanation score distances across multiple instances is computed using Equation~\ref{equ.Spearman}:

\begin{multline}
\label{equ.Spearman}
\rho = \text{Spearman's corr. } ( \{ D_A(M_a, M_b, x_i) \}_{i=1}^N, \\
\{ D_E(M_a, M_b, x_i) \}_{i=1}^N 
\end{multline} where \( N \) is the total number of inputs, and \( \rho \) measures the correlation between the attention-based reasoning and explanation scores.

A high \( \rho \) value signifies that models with similar attention weights also tend to have similar explanation patterns, suggesting a strong correlation between attention-based reasoning and explanation consistency. By employing this attention-weight-based approach, we can effectively evaluate whether models trained with different random seeds exhibit consistent reasoning mechanisms. This framework provides a robust method for assessing the reliability and robustness of explainability techniques, ensuring that models with similar attention distributions also maintain similar explanations, even when their final predictions might differ. This evaluation is crucial in understanding whether attention weights serve as a reliable indicator of reasoning similarity across models.

\subsubsection{Contrastivity} 
Contrastivity is a critical evaluation metric for assessing the effectiveness of XAI methods, particularly in classification tasks \cite{stepin2021survey}. It focuses on how well an XAI method can differentiate between different classes through its explanations, providing insight into why a model chooses one class over another. For example, to assess the difference between the two classes (positive or negative), we can compare the explanations for different class predictions and see if the explanations for one class are distinct from those for another. This practical use of contrastivity helps us to understand the effectiveness of XAI methods. We used Kullback-Leibler Divergence (KL Divergence) to quantify the contrastivity metric in feature importance distributions. KL Divergence is ideal for its sensitivity to differences in feature importance distributions and its focus on the direction of divergence, making it perfect for analyzing and comparing tokens/words (feature) importance across different classes \cite{kullback1951information} using Equation~\ref{contastivity}.
 
\begin{equation}
\label{contastivity}
    \text{KL}(P \parallel Q) = \sum_{i=1}^n P(i) \log \left( \frac{P(i)}{Q(i)} \right)
\end{equation}
where $P(i)$ and $Q(i)$ represent the importance of feature $i$ in one class and in a different class, respectively, and $n$ is the total number of tokens/words in the given instance. $P$ and $Q$ are the distribution of feature importance for the two different classes.

High contrastivity means that the XAI method effectively highlights different features for different classes.
The positive attributions should be associated with the target label, and the negative attributions should be associated with the opposite class.

\section{Experiments}  
\label{section:experiments}
We used a transformer-based LLM with varied levels of complexity to focus on text data and a classification task. To provide clear insights into our experiment, we included five distinct XAI methods, each representing five categories of XAI techniques. We then evaluated these methods using four specific metrics.

\subsection{Datasets}  
We used two distinct datasets for our study: IMDB \footnote{https://www.kaggle.com/datasets/columbine/imdb-dataset-sentiment-analysis-in-csv-format} Movie Reviews and Tweet Sentiment Extraction. The IMDB Movie Reviews dataset consists of 50,000 movie reviews, each labeled as either positive or negative. The Tweet Sentiment Extraction (TSE) \footnote{https://www.kaggle.com/c/tweet-sentiment-extraction} dataset consists of 31,016 tweets labeled with sentiments such as positive, negative, or neutral. Tweets are typically short texts. We randomly split each dataset into 80\% for training and 20\% for testing.

\subsection{Models} 
\begin{figure}[h]
\centering
\includegraphics[width=\linewidth]{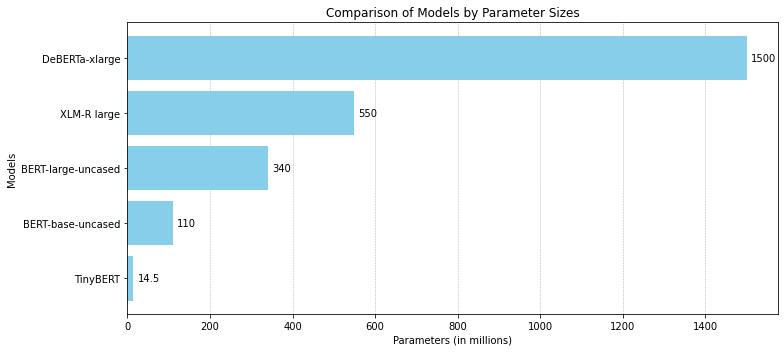}
\caption{The selected transformer-based models by complexity band parameter size.}
\label{fig:models}
\end{figure}

We conducted experiments using commonly used transformer-based models, including TinyBERT \cite{jiao2019tinybert}, BERT-base-uncased \cite{devlin2018bert}, BERT-large-uncased \cite{devlin2018bert}, XLM-R large \cite{conneau2019unsupervised}, and DeBERTa-xlarge \cite{he2020deberta}. These models were chosen primarily for their varying levels of complexity and parameter sizes. This baseline model selection enables a comprehensive comparison and analysis of explainability techniques across complex models \cite{yigezu2024ethio}. Figure ~\ref{fig:models}  illustrates the selected transformer-based models and their respective parameter sizes. By evaluating the effectiveness of explainability techniques, we aim to understand how model complexity and size influence the interpretability and transparency of these models in practical applications. 

\subsection{Representative XAI Techniques} 

We categorized explainability techniques into five groups: model simplification, perturbation-based methods, gradient-based methods, layer-wise relevance propagation, and attention mechanisms. We selected widely used techniques from each category. We employed them: LIME for model simplification, SHAP for perturbation-based, Integrated Gradients for gradient-based, LRP-$\epsilon$($epsilon$) for layer-wise relevance, and AMV for attention-based methods.

\begin{table*}[h!]
\caption{The quantitative results of the Human-reasoning Agreement metric on various XAI methods and LLM models for IMDB and TSE data. Higher scores indicate better agreement. }
\centering
\scriptsize
\begin{adjustbox}{max width=\textwidth}
\begin{tabular}{ccccccccccccccccccccccc}
\toprule
& \multicolumn{5}{c}{IMDB} & \multicolumn{5}{c}{TSE} \\
\cmidrule(lr){2-6} \cmidrule(lr){7-11}
 &TinyBERT  & BERTbase &  BERTlarge & XLM-R &  DeBERTa\_xlarge &TinyBERT & BERTbase & BERTlarge & XLM-R &  DeBERTa\_xlarge  \\
\midrule
LIME & \textbf{ 0.8774} & \textbf{0.6981 } &\textbf{0.8903}& \textbf{0.9445} & \textbf{0.9685} &\textbf{ 0.7566} & 0.4689 & \textbf{0.8023} & \textbf{0.8869} & \textbf{0.9118}  \\
SHAP & 0.4135 & 0.4354 &0.5012& 0.5634 & 0.6625 & 0.5231 & 0.5728 & 0.6002 & 0.6894 & 0.7254  \\
LRP & 0.6427 & 0.2736 & 0.2078 & 0.2011 & 0.1984 & 0.7223 & \textbf{0.5986 }& 0.5023 & 0.4533 & 0.3689 \\
 Integrated Gradients& 0.0697 & 0.0599 & 0.2178 & 0.2985 & 0.3721 & 0.1879 & 0.1001 & 0.3657 & 0.3844 & 0.4102 \\
AMV & 0.1658 & 0.1459 & 0.1001 & 0.0859 & 0.0653 & 0.2136 & 0.1759 & 0.1325 & 0.0962& 0.0593 \\
\bottomrule
\end{tabular}
\end{adjustbox}
\label{HA}
\end{table*}

\begin{table*}[h!]
\caption{The quantitative results of the Robustness metric on various XAI methods and LLM models for IMDB and TSE data. Lower scores indicate better robustness.}
\centering
\scriptsize
\begin{adjustbox}{max width=\textwidth}
\begin{tabular}{ccccccccccccccccccccccc}
\toprule
& \multicolumn{5}{c}{IMDB} & \multicolumn{5}{c}{TSE} \\
\cmidrule(lr){2-6} \cmidrule(lr){7-11}
 &TinyBERT  & BERTbase &  BERTlarge & XLM-R &  DeBERTa\_xlarge &TinyBERT & BERTbase & BERTlarge & XLM-R &  DeBERTa\_xlarge  \\
\midrule
LIME & 0.0056 & 0.0058 & 0.0061 & 0.0078 & 0.0092 & 0.0043 & 0.0049& 0.0053 & 0.0060 & 0.0068 \\
SHAP & 0.0356 & 0.0387 &0.1258 & 0.1547 & 0.2139 & 0.0258 & 0.0301& 0.03662& 0.1321 & 0.1965  \\
LRP & 0.3214 & 0.5431 & 0.7621 & 0.8549 & 0.9124 & 0.3392 & 0.4895 & 0.5684 & 0.6855 & 0.8215 \\
Integrated Gradients& 0.2108 & 0.3209 & 0.3987 & 0.3288 & 0.4411 & 0.0929 & 0.2716 & 0.2985 & 0.3284 & 0.4388 \\
AMV & \textbf{0.0020} & \textbf{0.0023} & \textbf{0.0056} & \textbf{0.0058} & \textbf{0.0073} & \textbf{0.0014} & \textbf{0.0019} & \textbf{0.0032} & \textbf{0.0041} & \textbf{0.0051} \\
\bottomrule
\end{tabular}
\end{adjustbox}
\label{robustness}
\end{table*}

\section{Results and Discussion} 
\label{section:results}
We present the results of each evaluation metric on various XAI methods and LLMs across the IMDB (long texts) and TSE (short texts) datasets, considering model complexity from TinyBERT (14.5 million parameters) to DeBERTa-large (1.5 billion parameters). The primary focus of this study is not on the performance of the models themselves, but on the effectiveness of the XAI methods on these LLMs.

\textbf{Human-reasoning Agreement (HA):}
Table \ref{HA} presents HA results for 100 randomly selected instances per dataset, using annotations from three ML experts as the baseline. For the IMDB dataset (longer texts), LIME consistently achieves high HA scores, improving with model size and reaching its peak (0.9685) with DeBERTa-xlarge. SHAP shows moderate performance but improves significantly for larger models. In contrast, LRP and AMV perform poorly, with a notable decline as model complexity increases. The integrated gradient shows moderate HA scores, which benefit slightly from larger models.
LIME remains the top performer for the TSE dataset (shorter texts), while SHAP adapts well and performs better than the IMDB dataset, especially on larger models. LRP and AMV show low performance across both datasets, struggling as model complexity increases. Integrated gradient improves with larger models but remains less aligned with human rationales than LIME and SHAP. LIME provides the strongest alignment across datasets and models.

 \begin{table*}[htbp!]
\caption{The quantitative results of Consistency metric on various XAI methods and LLM models for IMDB and TSE data. Higher scores indicate better consistency.}
\centering
\scriptsize
\begin{adjustbox}{max width=\textwidth}
\begin{tabular}{ccccccccccccccccccccccc}
\toprule
& \multicolumn{5}{c}{IMDB} & \multicolumn{5}{c}{TSE} \\
\cmidrule(lr){2-6} \cmidrule(lr){7-11}
 &TinyBERT  & BERTbase &  BERTlarge & XLM-R &  DeBERTa\_xlarge &TinyBERT & BERTbase & BERTlarge & XLM-R &  DeBERTa\_xlarge  \\
\midrule
LIME & 0.9665 & 0.9741 & 0.9800 & 0.9837 & 0.9895 & 0.7568& 0.8425 & 0.8962 & 0.9228& 0.9556 \\
SHAP & 0.9002 & 0.9368 &0.9487 & 0.9569 & 0.9775 & 0.8322 & 0.8598 & 0.8896& 0.9002& 0.09324  \\
LRP & 0.9417 & 0.9642 & 0.9754 & 0.9801& 0.9887 & 0.8556 & 0.8901 & 0.9223 & 0.9596 & 0.9713 \\
Integrated Gradients& 0.8799 & 0.9389 & 0.9435 & 0.9498 & 0.9629 & 0.7475 & 0.7699& 0.8012 & 0.8356& 0.8649 \\
AMV & \textbf{0.9999} & \textbf{0.9999} & \textbf{0.9999} & \textbf{0.9999} & \textbf{0.9999} & \textbf{0.9999} & \textbf{0.9999} & \textbf{0.9999} & \textbf{0.9999} & \textbf{0.9999} \\
\bottomrule
\end{tabular}
\end{adjustbox}
\label{consitency}
\end{table*}

\begin{table*}[htbp!]
\centering
\caption{The quantitative results of Contrastivity metric on various XAI methods and LLM models for IMDB and TSE data. Higher scores indicate better.}
\scriptsize
\begin{adjustbox}{max width=\textwidth}
\begin{tabular}{ccccccccccccccccccccccc}
\toprule
& \multicolumn{5}{c}{IMDB} & \multicolumn{5}{c}{TSE} \\
\cmidrule(lr){2-6} \cmidrule(lr){7-11}
 & TinyBERT & BERTbase & BERTlarge & XLM-R & DeBERTa\_xlarge & TinyBERT & BERTbase & BERTlarge & XLM-R & DeBERTa\_xlarge \\
\midrule
LIME & 0.7065 & \textbf{0.7226} & 0.6545 & 0.6288 & 0.5766 & 0.6863 & \textbf{0.7145} & 0.6631 & 0.6251 & 0.5838 \\
SHAP & 0.6449 & 0.6694 & 0.6208 & 0.5565 & 0.5448 & 0.6552 & 0.6727 & 0.6075 & 0.5709 & 0.5356 \\
LRP & \textbf{0.7723} & 0.5797 & \textbf{0.7958} & \textbf{0.8456} & \textbf{0.9371} & \textbf{0.8598} & 0.7126 & \textbf{0.8540} & \textbf{0.8797} & \textbf{0.9367} \\
Integrated Gradients& 0.7502 & 0.6155 & 0.4785 & 0.4366 & 0.3899 & 0.8011 & 0.7155 & 0.6099 & 0.5322 & 0.4141 \\
AMV & 0.1546 & 0.1168 & 0.1012 & 0.0761 & 0.0384 & 0.1841 & 0.1616 & 0.1243 & 0.1050 & 0.0680 \\
\bottomrule
\end{tabular}
\end{adjustbox}
\label{contrastivity}
\end{table*}

\textbf{Robustness:}
Table \ref{robustness} shows the robustness results across various models. LIME and AMV consistently exhibit high robustness on the IMDB and TSE datasets, with minimal variations in their scores regardless of text length or model complexity, indicating stable and reliable explanations. SHAP demonstrates moderate robustness, but its stability decreases significantly as model size increases. Integrated Gradient performs moderately well, showing more stability than LRP but less than LIME.  LRP is the least robust, with high variability and instability, especially for larger models.

\textbf{Consistency}:
Table \ref{consitency} presents the Consistency metric results on the IMDB (longer texts) and TSE (shorter texts) datasets. LIME consistently provides reliable explanations for both datasets, improving performance as model complexity increases. SHAP also demonstrates high consistency, particularly for larger models, while LRP and Integrated Gradients show stable performance, offering reliable explanations for more complex models. AMV achieves nearly perfect consistency across all models and datasets, delivering identical explanations regardless of model size or input text length. 
AMV and LIME are the most consistent techniques across IMDB and TSE, while SHAP, LRP, and Integrated Gradients show reasonable stability.

\begin{table*}[h!]
\caption{Combined Weighted-metrics Scores (CWS) for various XAI methods on different LLM models using IMDB and TSE datasets. Higher CWS values indicate better overall XAI performance.}
\centering
\scriptsize
\begin{adjustbox}{max width=\textwidth}
\begin{tabular}{ccccccccccccccccccccccc}
\toprule
& \multicolumn{5}{c}{IMDB} & \multicolumn{5}{c}{TSE} \\
\cmidrule(lr){2-6} \cmidrule(lr){7-11}
 & TinyBERT  & BERTbase &  BERTlarge & XLM-R & DeBERTa\_xlarge & TinyBERT & BERTbase & BERTlarge & XLM-R & DeBERTa\_xlarge  \\
\midrule
LIME & \textbf{0.8862} & \textbf{0.8755} & \textbf{0.8797} & \textbf{0.8873} & \textbf{0.8611} & \textbf{0.7989} & 0.7468 & \textbf{0.7785} & \textbf{0.8572} & \textbf{0.8611} \\
SHAP & 0.7308 & 0.7507 & 0.7427 & 0.7504 & 0.7427 & 0.7308 & \textbf{0.7507} & 0.7427 & 0.7504 & 0.7516 \\
LRP & 0.7588 & 0.5686 & 0.7329 & 0.6741 & 0.6809 & 0.6779 & 0.6562 & 0.6677 & 0.7055 & 0.7285 \\
Integrated Gradients & 0.6982 & 0.7199 & 0.7537 & 0.7571 & 0.7438 & 0.6907 & 0.7078 & 0.7226 & 0.7371 & 0.7526 \\
AMV & 0.5796 & 0.5241 & 0.5637 & 0.5561 & 0.5442 & 0.5991 & 0.5920 & 0.5733 & 0.5593 & 0.5436 \\
\bottomrule
\end{tabular}
\end{adjustbox}
\label{CombinedMetrics}
\end{table*}

\textbf{Contrastivity}: 
Table \ref{contrastivity} presents the results of the Contrastivity metric on the IMDB (long texts) and TSE (short texts) datasets. LRP is the most effective technique, particularly for complex models, showing a strong ability to highlight contrasting features. LIME performs well with smaller models, but its effectiveness decreases as model complexity increases. SHAP offers moderate contrastivity but also sees diminishing performance with more complex models.  Integrated Gradient performs well for simpler models but struggles as complexity increases, while AMV shows consistently poor contrastivity across both datasets and models, failing to reliably differentiate features in model predictions.

\textbf{Overall}, The study highlights that no single XAI technique excels universally across all metrics and models. However, our rigorous evaluation process has identified some reliable XAI techniques. LIME consistently performs well across multiple evaluation metrics, making it a reliable choice for generating explanations that align with human reasoning, robustness, and consistency. Despite its limitations in contrastivity, AMV excels in robustness and consistency, making it suitable for applications where stability and reliability are paramount. An LRP shows promise in contrastivity, particularly for complex models, indicating its potential for tasks requiring identifying contrasting features. SHAP and Integrated Gradient demonstrate moderate performance across all metrics and models. 


\textbf{Combined Weighted-metrics Scores (CWS)}: CWS evaluates XAI methods across LLMs using four metrics: Human-reasoning Agreement (HA), Robustness (R), Consistency (Cn), and Contrastivity (Ct). Higher scores are preferred for HA, Cn, and Ct, while lower scores (closer to 0) are ideal for R, indicating stability under perturbations. It is normalized to align R with other metrics as $(1 - R)$. All metrics are equally weighted ($\omega = 0.25$), but weights can be adjusted based on priority. CWS is computed using Equation \ref{cws}: 
\begin{equation}
\label{cws}
\text{CWS} = \omega_{HA} \cdot HA + \omega_{Cn} \cdot Cn + \omega_{Ct} \cdot Ct + \omega_{R} \cdot (1 - R),  
\end{equation}
where $\omega_{HA} + \omega_{Cn} + \omega_{Ct} + \omega_{R} = 1$.




In the CWS evaluation approach, LIME performs best across models, SHAP shows stable results, LRP is inconsistent, and Integrated Gradients shows moderate reliability, performing well on specific models but varying across datasets. AMV performs the lowest overall, as shown in Table \ref{CombinedMetrics}. Appendix \ref{sec:appendix} provides the visualization of all results.

\section{Conclusion}  \label{sectionConclusion}
We presented a comprehensive framework to evaluate XAI techniques for LLMs using four key metrics: Human-Reasoning Agreement, Robustness, Consistency, and Contrastivity. This systematic evaluation across diverse models and datasets provides valuable insights into each technique’s alignment with human judgment, stability under perturbations, and ability to highlight distinct features. LIME achieved the highest overall scores despite its computational cost, AMV excelled in Robustness and Consistency, while LRP was the most effective for Contrastivity in complex models. Integrated Gradient demonstrated balanced performance across all metrics.

Our study provides a structured and adaptable approach for assessing XAI methods across different models, tasks, and datasets. This flexible framework offers a scalable solution for real-world applications, with insights from our analysis of LIME, AMV, LRP, and Integrated Gradients guiding the selection of appropriate XAI techniques based on specific priorities such as robustness and alignment with humans.

\section{Limitations and Future Directions} 
Our study is limited to specific metrics, models, XAI techniques, and classification tasks. Future work can expand by incorporating additional XAI techniques, models, under-resourced languages, and downstream tasks to refine the metrics and enhance real-world applicability. The study does not account for the computational complexity of XAI techniques; addressing this will be crucial for improving the scalability and practicality of our evaluation framework in large-scale applications.

\bibliographystyle{elsarticle-num}
\bibliography{bib.bib}

\begin{thebibliography}{10}
\expandafter\ifx\csname url\endcsname\relax
  \def\url#1{\texttt{#1}}\fi
\expandafter\ifx\csname urlprefix\endcsname\relax\def\urlprefix{URL }\fi
\expandafter\ifx\csname href\endcsname\relax
  \def\href#1#2{#2} \def\path#1{#1}\fi

\bibitem{radford2018improving}
A.~Radford, K.~Narasimhan, T.~Salimans, I.~Sutskever, et~al., Improving language understanding by generative pre-training (2018).

\bibitem{shakil2024evaluating}
H.~Shakil, A.~M. Mahi, P.~Nguyen, Z.~Ortiz, M.~T. Mardini, Evaluating text summaries generated by large language models using openai's gpt, arXiv preprint arXiv:2405.04053 (2024).

\bibitem{devlin2018bert}
J.~Devlin, M.-W. Chang, K.~Lee, K.~Toutanova, Bert: Pre-training of deep bidirectional transformers for language understanding, arXiv preprint arXiv:1810.04805 (2018).

\bibitem{mersha2024semantic}
M.~A. Mersha, J.~Kalita, et~al., Semantic-driven topic modeling using transformer-based embeddings and clustering algorithms, Procedia Computer Science 244 (2024) 121--132.

\bibitem{das2020opportunities}
A.~Das, P.~Rad, Opportunities and challenges in explainable artificial intelligence (xai): A survey, arXiv preprint arXiv:2006.11371 (2020).

\bibitem{arrieta2020explainable}
A.~B. Arrieta, N.~D{\'\i}az-Rodr{\'\i}guez, J.~Del~Ser, A.~Bennetot, S.~Tabik, A.~Barbado, S.~Garc{\'\i}a, S.~Gil-L{\'o}pez, D.~Molina, R.~Benjamins, et~al., Explainable artificial intelligence (xai): Concepts, taxonomies, opportunities and challenges toward responsible ai, Information fusion 58 (2020) 82--115.

\bibitem{regulation2016regulation}
P.~Regulation, Regulation (eu) 2016/679 of the european parliament and of the council, Regulation (eu) 679 (2016) 2016.

\bibitem{langer2021we}
M.~Langer, D.~Oster, T.~Speith, H.~Hermanns, L.~K{\"a}stner, E.~Schmidt, A.~Sesing, K.~Baum, What do we want from explainable artificial intelligence (xai)?--a stakeholder perspective on xai and a conceptual model guiding interdisciplinary xai research, Artificial Intelligence 296 (2021) 103473.

\bibitem{shah2021neural}
V.~Shah, S.~R. Konda, Neural networks and explainable ai: Bridging the gap between models and interpretability, INTERNATIONAL JOURNAL OF COMPUTER SCIENCE AND TECHNOLOGY 5~(2) (2021) 163--176.

\bibitem{mersha2024explainability}
M.~Mersha, M.~Bitewa, T.~Abay, J.~Kalita, Explainability in neural networks for natural language processing tasks, arXiv preprint arXiv:2412.18036 (2024).

\bibitem{ribeiro2016should}
M.~T. Ribeiro, S.~Singh, C.~Guestrin, " why should i trust you?" explaining the predictions of any classifier, in: Proceedings of the 22nd ACM SIGKDD international conference on knowledge discovery and data mining, 2016, pp. 1135--1144.

\bibitem{zeiler2014visualizing}
M.~D. Zeiler, R.~Fergus, Visualizing and understanding convolutional networks, in: Computer Vision--ECCV 2014: 13th European Conference, Zurich, Switzerland, September 6-12, 2014, Proceedings, Part I 13, Springer, 2014, pp. 818--833.

\bibitem{sundararajan2017axiomatic}
M.~Sundararajan, A.~Taly, Q.~Yan, Axiomatic attribution for deep networks, in: International conference on machine learning, PMLR, 2017, pp. 3319--3328.

\bibitem{bach2015pixel}
S.~Bach, A.~Binder, G.~Montavon, F.~Klauschen, K.-R. M{\"u}ller, W.~Samek, On pixel-wise explanations for non-linear classifier decisions by layer-wise relevance propagation, PloS one 10~(7) (2015) e0130140.

\bibitem{honnibal2017spacy}
M.~Honnibal, I.~Montani, spacy 2: Natural language understanding with bloom embeddings, convolutional neural networks and incremental parsing, To appear 7~(1) (2017) 411--420.

\bibitem{hassija2024interpreting}
V.~Hassija, V.~Chamola, A.~Mahapatra, A.~Singal, D.~Goel, K.~Huang, S.~Scardapane, I.~Spinelli, M.~Mahmud, A.~Hussain, Interpreting black-box models: a review on explainable artificial intelligence, Cognitive Computation 16~(1) (2024) 45--74.

\bibitem{pawlicki2024evaluating}
M.~Pawlicki, A.~Pawlicka, F.~Uccello, S.~Szelest, S.~D’Antonio, R.~Kozik, M.~Chora{\'s}, Evaluating the necessity of the multiple metrics for assessing explainable ai: A critical examination, Neurocomputing (2024) 128282.

\bibitem{atanasova2024diagnostic}
P.~Atanasova, A diagnostic study of explainability techniques for text classification, in: Accountable and Explainable Methods for Complex Reasoning over Text, Springer, 2024, pp. 155--187.

\bibitem{mersha2024explainable}
M.~Mersha, K.~Lam, J.~Wood, A.~AlShami, J.~Kalita, Explainable artificial intelligence: A survey of needs, techniques, applications, and future direction, Neurocomputing (2024) 128111.

\bibitem{shapley1953value}
L.~S. Shapley, et~al., A value for n-person games (1953).

\bibitem{montavon2017explaining}
G.~Montavon, S.~Lapuschkin, A.~Binder, W.~Samek, K.-R. M{\"u}ller, Explaining nonlinear classification decisions with deep taylor decomposition, Pattern recognition 65 (2017) 211--222.

\bibitem{chefer2021transformer}
H.~Chefer, S.~Gur, L.~Wolf, Transformer interpretability beyond attention visualization, in: Proceedings of the IEEE/CVF conference on computer vision and pattern recognition, 2021, pp. 782--791.

\bibitem{ali2022xai}
A.~Ali, T.~Schnake, O.~Eberle, G.~Montavon, K.-R. M{\"u}ller, L.~Wolf, Xai for transformers: Better explanations through conservative propagation, in: International Conference on Machine Learning, PMLR, 2022, pp. 435--451.

\bibitem{karras2020analyzing}
T.~Karras, S.~Laine, M.~Aittala, J.~Hellsten, J.~Lehtinen, T.~Aila, Analyzing and improving the image quality of stylegan, in: Proceedings of the IEEE/CVF conference on computer vision and pattern recognition, 2020, pp. 8110--8119.

\bibitem{kindermans2019reliability}
P.-J. Kindermans, S.~Hooker, J.~Adebayo, M.~Alber, K.~T. Sch{\"u}tt, S.~D{\"a}hne, D.~Erhan, B.~Kim, The (un) reliability of saliency methods, Explainable AI: Interpreting, explaining and visualizing deep learning (2019) 267--280.

\bibitem{jacovi2020towards}
A.~Jacovi, Y.~Goldberg, Towards faithfully interpretable nlp systems: How should we define and evaluate faithfulness?, arXiv preprint arXiv:2004.03685 (2020).

\bibitem{bansal2021does}
G.~Bansal, T.~Wu, J.~Zhou, R.~Fok, B.~Nushi, E.~Kamar, M.~T. Ribeiro, D.~Weld, Does the whole exceed its parts? the effect of ai explanations on complementary team performance, in: Proceedings of the 2021 CHI conference on human factors in computing systems, 2021, pp. 1--16.

\bibitem{arras2022clevr}
L.~Arras, A.~Osman, W.~Samek, Clevr-xai: A benchmark dataset for the ground truth evaluation of neural network explanations, Information Fusion 81 (2022) 14--40.

\bibitem{deyoung2019eraser}
J.~DeYoung, S.~Jain, N.~F. Rajani, E.~Lehman, C.~Xiong, R.~Socher, B.~C. Wallace, Eraser: A benchmark to evaluate rationalized nlp models, arXiv preprint arXiv:1911.03429 (2019).

\bibitem{arreche2024xai}
O.~Arreche, T.~R. Guntur, J.~W. Roberts, M.~Abdallah, E-xai: Evaluating black-box explainable ai frameworks for network intrusion detection, IEEE Access (2024).

\bibitem{mersha2024ethio}
M.~A. Mersha, G.~Y. Bade, J.~Kalita, O.~Kolesnikova, A.~Gelbukh, et~al., Ethio-fake: Cutting-edge approaches to combat fake news in under-resourced languages using explainable ai, Procedia Computer Science 244 (2024) 133--142.

\bibitem{shakil2024utilizing}
H.~Shakil, Z.~Ortiz, G.~C. Forbes, J.~Kalita, Utilizing gpt to enhance text summarization: A strategy to minimize hallucinations, Procedia Computer Science 244 (2024) 238--247.

\bibitem{guan2019towards}
C.~Guan, X.~Wang, Q.~Zhang, R.~Chen, D.~He, X.~Xie, Towards a deep and unified understanding of deep neural models in nlp, in: International conference on machine learning, PMLR, 2019, pp. 2454--2463.

\bibitem{shakil2024abstractive}
H.~Shakil, A.~Farooq, J.~Kalita, Abstractive text summarization: State of the art, challenges, and improvements, Neurocomputing (2024) 128255.

\bibitem{tonja2023first}
A.~L. Tonja, M.~Mersha, A.~Kalita, O.~Kolesnikova, J.~Kalita, First attempt at building parallel corpora for machine translation of northeast india's very low-resource languages, arXiv preprint arXiv:2312.04764 (2023).

\bibitem{mersha2025evaluating}
M.~A. Mersha, M.~G. Yigezu, J.~Kalita, Evaluating the effectiveness of xai techniques for encoder-based language models, Knowledge-Based Systems (2025) 113042.

\bibitem{nauta2023anecdotal}
M.~Nauta, J.~Trienes, S.~Pathak, E.~Nguyen, M.~Peters, Y.~Schmitt, J.~Schl{\"o}tterer, M.~van Keulen, C.~Seifert, From anecdotal evidence to quantitative evaluation methods: A systematic review on evaluating explainable ai, ACM Computing Surveys 55~(13s) (2023) 1--42.

\bibitem{rosenfeld2021better}
A.~Rosenfeld, Better metrics for evaluating explainable artificial intelligence (2021).

\bibitem{nogueira2018stability}
S.~Nogueira, K.~Sechidis, G.~Brown, On the stability of feature selection algorithms, Journal of Machine Learning Research 18~(174) (2018) 1--54.

\bibitem{huang2022conceptexplainer}
J.~Huang, A.~Mishra, B.~C. Kwon, C.~Bryan, Conceptexplainer: Interactive explanation for deep neural networks from a concept perspective, IEEE Transactions on Visualization and Computer Graphics 29~(1) (2022) 831--841.

\bibitem{stepin2021survey}
I.~Stepin, J.~M. Alonso, A.~Catala, M.~Pereira-Fari{\~n}a, A survey of contrastive and counterfactual explanation generation methods for explainable artificial intelligence, IEEE Access 9 (2021) 11974--12001.

\bibitem{kullback1951information}
S.~Kullback, R.~A. Leibler, On information and sufficiency, The annals of mathematical statistics 22~(1) (1951) 79--86.

\bibitem{jiao2019tinybert}
X.~Jiao, Y.~Yin, L.~Shang, X.~Jiang, X.~Chen, L.~Li, F.~Wang, Q.~Liu, Tinybert: Distilling bert for natural language understanding, arXiv preprint arXiv:1909.10351 (2019).

\bibitem{conneau2019unsupervised}
A.~Conneau, K.~Khandelwal, N.~Goyal, V.~Chaudhary, G.~Wenzek, F.~Guzm{\'a}n, E.~Grave, M.~Ott, L.~Zettlemoyer, V.~Stoyanov, Unsupervised cross-lingual representation learning at scale, arXiv preprint arXiv:1911.02116 (2019).

\bibitem{he2020deberta}
P.~He, X.~Liu, J.~Gao, W.~Chen, Deberta: Decoding-enhanced bert with disentangled attention, arXiv preprint arXiv:2006.03654 (2020).

\bibitem{yigezu2024ethio}
M.~G. Yigezu, M.~A. Mersha, G.~Y. Bade, J.~Kalita, O.~Kolesnikova, A.~Gelbukh, Ethio-fake: Cutting-edge approaches to combat fake news in under-resourced languages using explainable ai, arXiv preprint arXiv:2410.02609 (2024).

\end{thebibliography}

\appendix

\section{Appendix}
\label{sec:appendix}

\begin{figure}[h] \vspace{-3em}
\centering
\includegraphics[width=\linewidth]{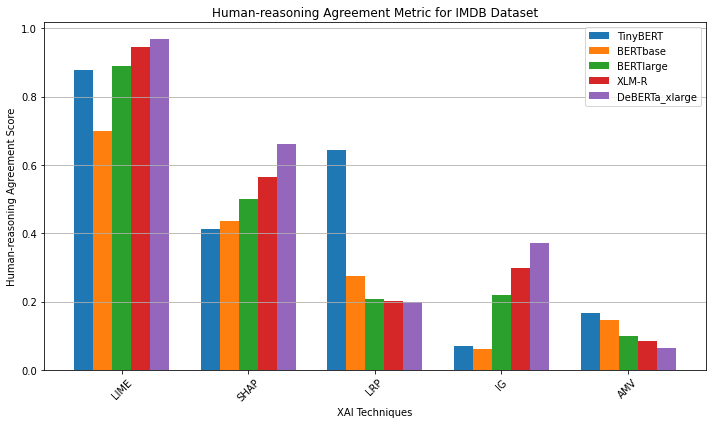}
\caption{ Human Agreement metric Scores for various XAI methods on different LLM models using IMDB dataset. Higher score values indicate better XAI performance.}
\end{figure}

\begin{figure}[h] \vspace{-3em}
\centering
\includegraphics[width=\linewidth]{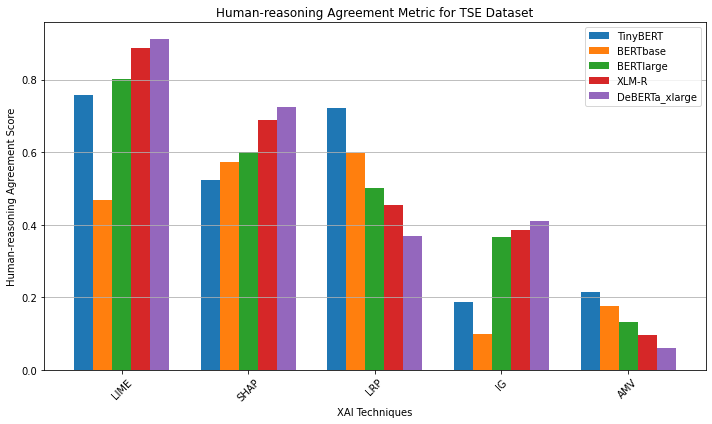}
\caption{ Human Agreement metric Scores for various XAI methods on different LLM models using TSE dataset. Higher score values indicate better XAI performance.}
\end{figure}

\begin{figure}[h]
\centering
\includegraphics[width=\linewidth]{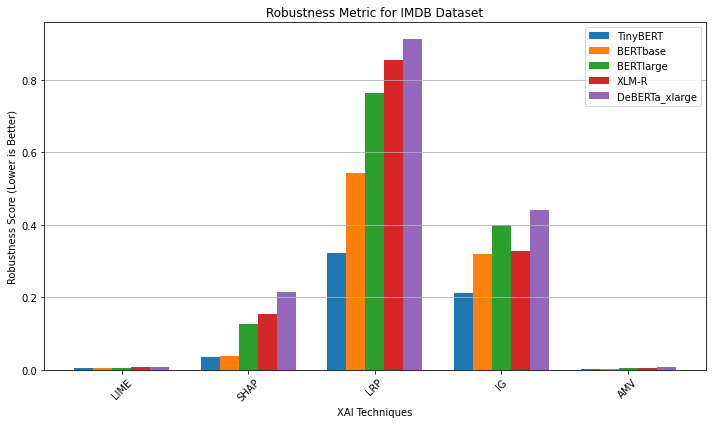}
\caption{ Robustness Metric for various XAI methods across different LLM models on the IMDB datasets. Lower Robustness scores indicate better XAI performance.}
\end{figure}

\begin{figure}[h]
\centering
\includegraphics[width=\linewidth]{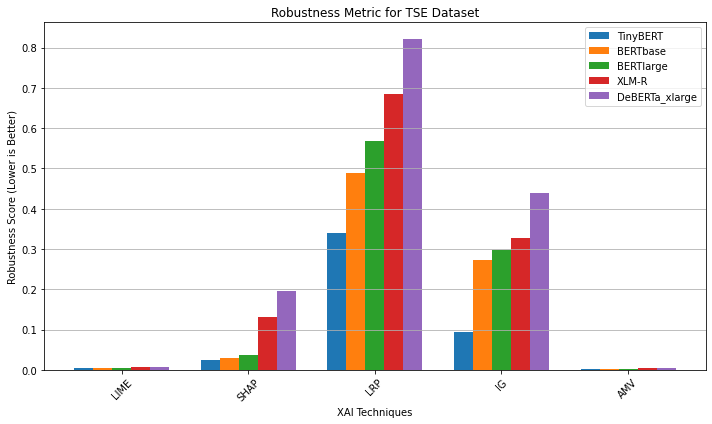}
\caption{ Robustness Metric for various XAI methods across different LLM models on the TSE datasets. Lower Robustness scores indicate better XAI performance.}
\end{figure}

\begin{figure}[h]
\centering
\includegraphics[width=\linewidth]{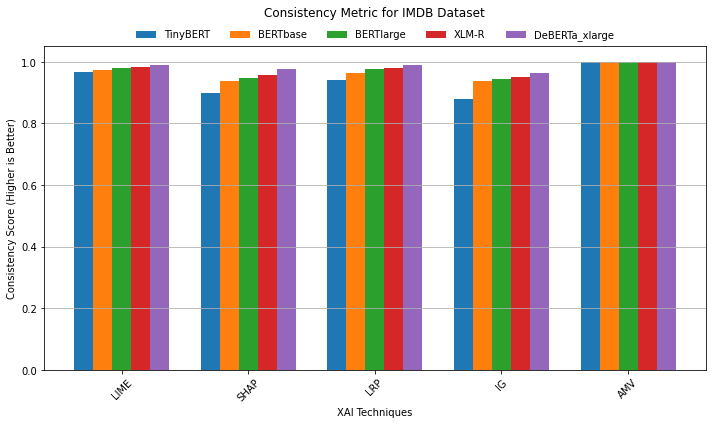}
\caption{ Consistency Metric for various XAI methods across different LLM models on the IMDB datasets. Higher score values indicate better XAI performance.}
\end{figure}
\begin{figure}[h]
\centering
\includegraphics[width=\linewidth]{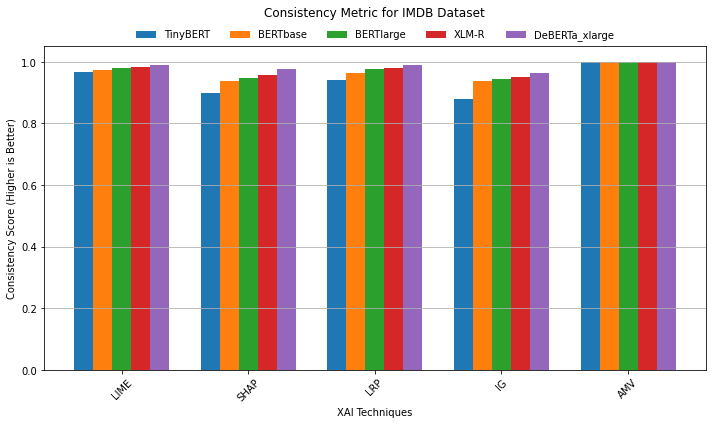}
\caption{ Consistency Metric for various XAI methods across different LLM models on the TSE datasets. Higher score values indicate better XAI performance.}
\end{figure}
\begin{figure}[h]
\centering
\includegraphics[width=\linewidth]{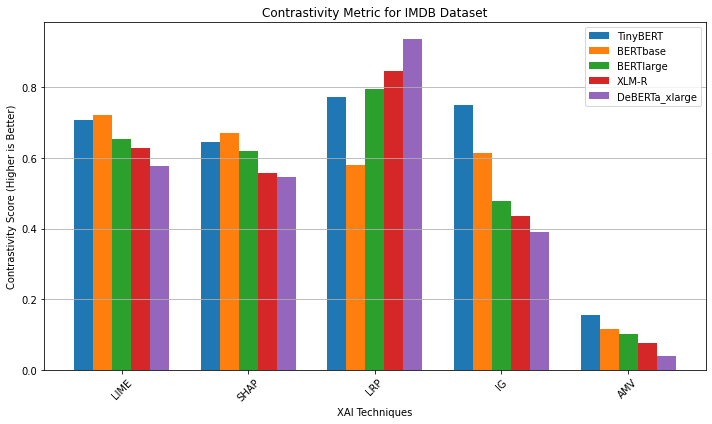}
\caption{ Contrastivity Metric for various XAI methods across different LLM models on the IMDB datasets. Higher score values indicate better XAI performance.}
\end{figure}
\begin{figure}[h]
\centering
\includegraphics[width=\linewidth]{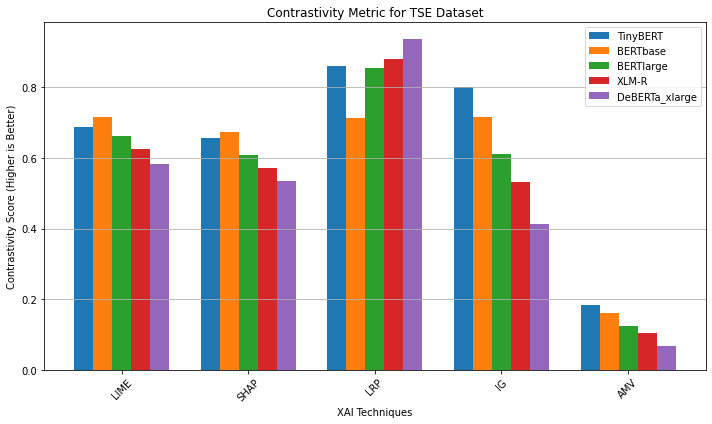}
\caption{ Contrastivity Metric for various XAI methods across different LLM models on the TSE datasets. Higher score values indicate better XAI performance}
\end{figure}
\begin{figure}[h]
\centering
\includegraphics[width=\linewidth]{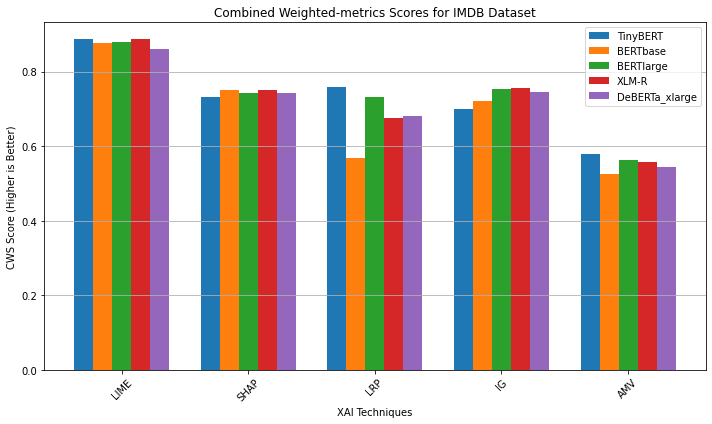}
\caption{ Combined Weighted-metrics Scores (CWS) for various XAI methods on different LLM models using IMDB datasets. Higher CWS values indicate better overall XAI performance.}
\end{figure}
\begin{figure}[h]
\centering
\includegraphics[width=\linewidth]{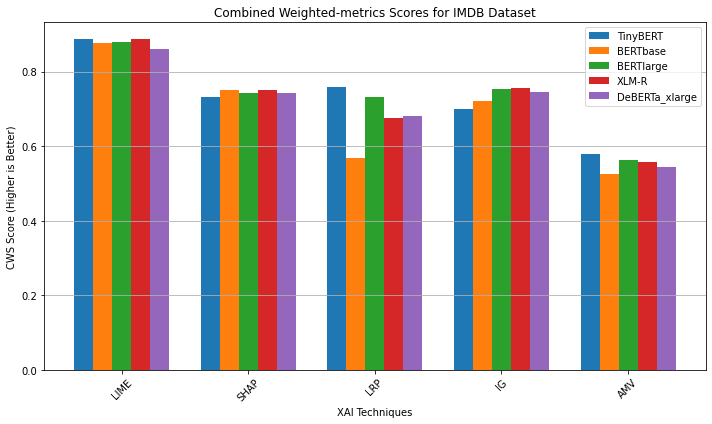}
\caption{ Combined Weighted-metrics Scores (CWS) for various XAI methods on different LLM models using TSE datasets. Higher CWS values indicate better overall XAI performance.}
\end{figure}

\end{document}